# Emotionally Expressive Robots: Implications for Children's Behavior toward Robot

Elisabetta Zibetti[1], Sureya Waheed Palmer[1], Rebecca Stower[1,2], Salvatore M. Anzalone[1]

*Abstract*— The growing development of robots with artificial emotional expressiveness raises important questions about their persuasive potential in children's behavior. While research highlights the pragmatic value of emotional expressiveness in human social communication, the extent to which robotic expressiveness can or should influence empathic responses in children is grounds for debate. In a pilot study with 22 children (aged 7-11) we begin to explore the ways in which different levels of embodied expressiveness (body only, face only, body and face) of two basic emotions (happiness and sadness) displayed by an anthropomorphic robot (QTRobot) might modify children's behavior in a child-robot cooperative turn-taking game. We observed that children aligned their behavior to the robot's inferred emotional state. However, higher levels of expressiveness did not result in increased alignment. The preliminary results reported here provide a starting point for reflecting on robotic expressiveness and its role in shaping children's social-emotional behavior toward robots as social peers in the near future.

## I. INTRODUCTION

In an era where technology increasingly blurs the boundaries between humans and machines [1], expressive anthropomorphic robots are of particular interest for examining the social dimensions of child-robot interactions (CRI). Unlike other technologies, anthropomorphic robots are both physically present and socially evocative allowing them to interact with humans on "human terms". According to recent AI developments, robots will likely evolve into more autonomous entities equipped with cognitive skills as well as with social-emotional capabilities. As a result, they are expected to play a significant role as children's companions and develop increasingly sophisticated interactions with them on a social-emotional level. Examples are emotionally expressive robots, some, known as *empathic agents* which integrate AI-driven systems to detect children's emotions and respond with adaptive, empathy-mimicking feedback [2]. The underlying motivation is to enhance the fluidity and naturalness of interactions by making robots more attuned and responsive to human emotions. However, one could ask: how attuned and responsive are children themselves to the emotions expressed by anthropomorphic robots?

Understanding the social-emotional mechanisms governing child-robot relationship formation and the impact of their expressiveness across different levels of embodiment [3] as a means of influence and persuasion is essential [4] but it is still a relatively unexplored question [5]. Several studies have been conducted so far on expressive robot's persuasive behaviors. However this line of work inherently raises several ethical and developmental considerations [6], particularly when robots attempt to influence children's decision making [7] and behaviours in accord with their embodied artificial expressiveness. Grounded on the idea that emotion recognition among humans can trigger a process of persuasion [8] and empathic responses in 7-12 years old children with their peers [9], we put forward the hypothesis that robots endowed with an increased level of embodiment (body, facial, and combined expressions of emotions) are more likely to create a strong sense of social presence and make their emotional messages more impactful. To do so, we explore the use of nonverbal facial and bodily expressive cues in anthropomorphic robots, to investigate their role as interactive peers — that is, as 'social agents' — in a playful child-robot cooperation setting. This approach should allow us to critically anticipate their role as novel expressive embodied AI within the fabric of future societies. For this purpose, even in the absence of advanced AI-driven emotion recognition and adaptability [2], we consider that certain anthropomorphic expressive robots, such as Nao and QTRobot, remain valuable tools for fostering verbal and physical interactions with children on a social-emotional level [5]. Gaining better insight into whether children react "emphatically" to emotions expressed by robots and if so, how these emotions may encourage — or fail to encourage — "empathically" aligned behaviors toward them, in interactive scenarios similar to those that they can encounter when playing with other children.

## II. RELATED WORKS

### A. Anthropomorphic robots and children

Anthropomorphism is a central factor in mediating children's behavior towards robots [10]. Anthropomorphic robots can use nonverbal communication channels — facial expressions, gaze direction, posture, head and body movements — to engage children socially [11], [12]. Children aged 2 to 12 tend to spontaneously attribute a wide range of skills and cognitive abilities to anthropomorphic robots, just as they do to living beings [13]. This includes the attribution of mental states and emotions [14] or attributing agency and intent to their actions [15]. Children's communication with robots is partly shaped by this projection phenomena, making it similar to how they interact with other children. Children aged between 2 and 12 years old, readily interpret and react to multimodal emotional cues from robots, recognizing emotions and inferring internal states based on external cues, as facial configurations and body postures [16]. Younger children, in particular, tend to recognise social cues more easily and more accurately, leading to a heightened sense of a robot's sociality relative to older children. This difference in perception suggests that, while younger children may respond positively to simple social cues, older children are more capable of discerning robots' social attributes [17]. Thus, by leveraging human-like social cues and multimodal nonverbal communication, expressive emotional robots may be particularly effective in influencing younger children's behaviour.

[1] E. Zibetti, S. Palmer ; S.M Anzalone are at Laboratoire CHArt, Université Paris 8, Saint-Denis, FRANCE. e-mail: ezibetti@univ-paris8.fr (corresponding author); sureyaw@gmail.com, sanzalone@univ-paris8.fr.

[2] R. Stower was at Laboratoire CHArt. She is now at Ericsson becstower@gmail.com

*A. Expressiveness for Influence*

Emotion is widely recognised for its strong pragmatic and persuasive value. In real social interactions, nonverbal emotional expressions are not merely reflections of internal states to be perceived and recognized, but function as powerful interpersonal signals [8] that shape behaviours of the receivers according to a communicative persuasive purpose. Persuasion is generally defined as "an attempt to shape, reinforce, or change behaviors, feelings, or thoughts about an issue, object, or action" [18, p. 225]. It is a fundamental social process that facilitates attitude change, social influence, and cooperation [8]. Different emotion expressions elicit distinct persuasive effects, leading to varying outcomes in guiding decision-making and behavior [8] ultimately shaping social dynamics and enhancing compliance [19]. Emotions like happiness and sadness exert different communicative effects. For instance, the attribution of disappointment, inferred through a visual expression of sadness, can signal a need for assistance and enhance compliance with requests of a partner i.e., influencing the decision to offer support [20][1].

Expressiveness in HRI refers to a robotic system's ability to convey social and affective information through nonverbal modalities to enhance understanding, empathy, as well as the overall effectiveness of interactions. Expressiveness encompasses both the emotional and the social dimension [5], referring to the ability to accurately communicate emotions, attitudes, and interpersonal orientation cues. Social expressiveness, on the other hand, involves broader verbal and nonverbal expressive skills, as well as the capacity to engage and involve interaction partners effectively. In HRI, by using nonverbal expressive cues, robots can provide a more natural and intuitive means of communication to users, especially in terms of emotion [21]. Similar to H-H interactions, non-verbal cues of robots in HRI have been shown to possess a strong *persuasive value*, eliciting emotional, cognitive, and behavioral changes in humans (see [4] for a survey).

The potential of robots to persuade users through verbal and nonverbal cues has been a key area of interest that is increasingly explored [22], [23]. Early studies highlighted that participant complied significantly more with the robot's suggestions when it employed nonverbal cues than when it did not. Moreover, bodily cues were found to be more effective in persuasion than vocal cues alone [24]. In line with social agency theory [25], it has been observed that an increasing presence of social cues appears to elicit more social responsiveness to a robot by humans [26], [27]. That is, the more a robot displays social cues the more likely adults are to respond socially to its persuasive attempts [23]. Using an expressive robot head, Kühnlenz et al. [28] found that the robot could persuade adults to show increased helpfulness towards it. Finally, displays of affect in robots (such as distress) have been shown to dissuade adults from issuing unethical commands and influence them toward more ethical outcomes (e.g., [29]).

*B. Persuasive value of expressiveness in cooperation settings*

In adults, the effects of nonverbal communication (e.g., cues of expressions of happiness and sadness) have been tested on a variety of prosocial contexts, including helping a stranger and cooperating in economic games [30], [31]. In a cooperative game (i.e., the Prisoner's Dilemma), Takahashi et al. [30] demonstrated that a humanoid robot, even without sophisticated facial expressions, was nevertheless able to establish cooperative relationships with humans by using full-body movements. However, contradictory results exist. For instance, Becker et al. [31] used robotic gestures to enhance expressiveness and emotional engagement in an interactive HRI cooperative game (the Coin Entrustment Cooperation Game). In their study, an emotional robot (as opposed to a neutral robot) was associated with a reduction in trust and cooperation.

In contrast, studies conducted specifically on children show that children are inclined to follow robots' suggestions with greater attentiveness than adults [7], [32]. Children generally change their behavior in response to inferred robot intentions conveyed through non-verbal expressions of disappointment [33] or based on the robot prosocial behavior that is witnessed in a free interaction scenario [34]. However, studies showing that a robot's *emotion* can influence children's decision-making toward a prosocial outcome are still rare (e.g., see [4]), and empirical evidence in the literature remains limited. In addition, the persuasive value of robot expressions of affect (body and face emotions) in cooperation or competition games has not yet been addressed.

Expressive-based HRI relies on a continuous interplay between emotion recognition and responses based on inferred meaning. "*Empathetic" reactions* are only triggered when the observer perceives the emotional cue and interprets it within a specific contextual framework. While robot multimodal expressions may be clear and legible, their interpretation can be ambiguous and shaped by the context: as an example, a smile in a cooperative versus competitive setting could signal different, opposite intentions [31]. However, the contextual framework (e.g., cooperation or competition) in CRI studies is rarely considered a key factor in shaping appropriate robot persuasive value [35], [36]. Leite et al. [35] and Tielman et al. [36] remain key studies in showing how, in structured CRI, expressive verbal and non-verbal robots cues impact trust, engagement, and learning potential. For instance, in turn-taking competitive games like chess with an iCat robot [35] or quiz games [36], expressive anthropomorphic robots shape player responses in guiding their behavior.

---

[1] Compliance, is "a particular kind of response — acquiescence — to a particular kind of communication — a request … [where] the target recognises that he or she is being urged to respond in a desired way" [19, p. 592]. Nevertheless, compliance is different to conformity, which is "the act of changing one's behavior to match the responses of others" [19, p. 606].

Finally, despite ongoing efforts to develop multimodal expressive robots and the ongoing ethical debate about using expressive robots to elicit empathic responses [6] there is still a lack of studies about *how*, *if* and *in which situations* they can persuade children by positively influencing their behaviour based on their expressiveness and its increased level of embodiment [3].

## III. THE STUDY

This pilot study aims to investigate children's *alignment* in an original turn-taking CRI collaborative task. *Alignment* is equated with the concepts of conformity and compliance behavior [19], defined as the act of changing one's behavior to adapt to the robot's request, inferred by the child from the robot's expression. By assessing the persuasive potential of an expressive robot with increasing levels of expressive embodiment (Body Only- BO, Face Only- FO, Body and Face BF) of two basic emotions (happiness and sandiness) we hypothesise that:

**H1:** Children's actions will *align* with the robot's inferred internal state. When the robot expresses happiness, children will confirm their initial choice. When the robot expresses sadness, children will discard their initial choice and adjust their decision to align with the robot's inferred internal state.

**H2:** Higher level of embodiment in robot expressiveness will increase children's *alignment* with the robot's inferred internal state. We predict an increasing alignment according to the level of embodiment (modality BO = FO < BF).

### A. Design and Procedure

This preliminary experiment was conducted to examine the persuasive impact of the embodiment (BO, FO, BF) of two basic emotions (happiness and sadness) displayed by QTRobot during a turn-taking CRI guided co-construction task. In *the Necklace Task*, children are asked to design a necklace in cooperation with the QTRobot (Fig. 1). They are required to select coloured beads and present them to the robot. The robot only reacts "emotionally" to the child's bead choices and the child is expected to interpret the non-verbal robot feedback. Within the shared goal framework, children did not merely have to discriminate among the robot's expression but they have to actively infer the robot's internal state (i.e., its satisfied or disappointed by the child's choice) expressed by an emotion (happy or sad) performed with an increased level of expressiveness (BO, FO, BF). While interacting with the robot, children are left totally free to react as they wish to the robot's feedback and then autonomously decide if they want to keep or discard their choice. In this scenario, the cooperative design of the necklace entirely relies on children's prosocial behavior, triggering their natural helpfulness [33]. Rather than reflecting strict joint coordination, the task is designed to simulate a form of asymmetric cooperation where children retain full decision-making autonomy but are exposed to evaluative emotional feedback without relying on explicit verbal prompts. Although coordination between the agents (child and robot) is limited, the mutual consideration of actions (bead choice, robot's emotional feedback, and the child's response) positions this interaction within a minimal cooperative framework, this asymmetrical design draws inspiration from prior CRI studies (e.g., [30], [35]) where minimal cooperation is sufficient to examine the influence of emotions. The experiment followed a 2×3×2 within-subjects design, where each participant experienced the two emotions (Happiness and Sadness) across three levels of expressiveness (Body Only (BO) and Face Only (FO); Face and Body (FB) - Fig. 2), with two trials per condition, resulting in a total of 12 trials per participant. The order of the trials was semi-randomised. The study focused on happiness and sadness, as primary emotions tend to elicit stronger empathic responses compared to secondary emotions (e.g., jealousy) [37].

### B. Participants

Twenty-two ($N = 22$) children, (eight boys and fourteen girls) between the ages of 6 and 11 ($M_{age} = 8.36$, $SD = 1.43$), took part in the experiment. They were recruited for voluntary participation from one school in Paris. The age of the population was chosen in accordance with the state of the art on children's emotion recognition. By the age of 5, children attribute affective characteristics to robots [38] and can recognise basic emotion expressions such as happiness and sadness expressed by humanoid robots. Around the age of 7, children develop logical thinking and between 8 and 10, they begin to perceive moral behaviors, such as helping, as inherently "right" [39] and can follow cooperative game instructions.

Parental consent and verbal assent from the children were obtained, and ethical approval was granted by Paris 8 University (IRB CE-P8-2022-01-4). As a token of appreciation, children received sticker sheets to build a robot at home.

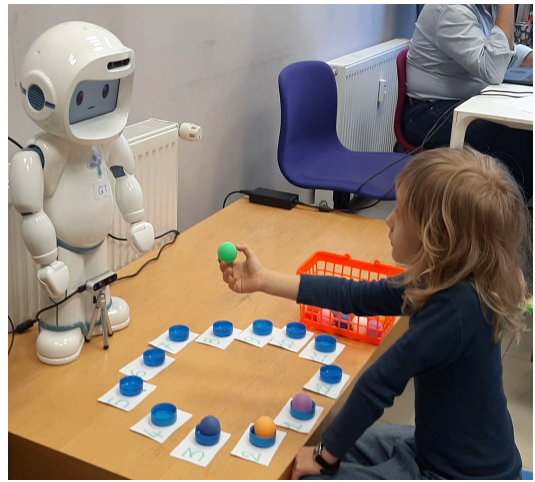

Figure 1. Experimental set-up with a participant and QTRobot.

### C. The Robot

The QT robot (LuxAI) was chosen due to its proved suitability for emotion recognition in children with ASD. Standing 63 cm tall and weighing 5 kg, it features a screen-based face capable of displaying six facial expressions in real-time, including happiness and sadness. Its motorised head and arms, with 14 degrees of freedom, enable natural upper-body movements. The robot was remotely controlled via Robot Operating Software (ROS) and Python in a Wizard of Oz setup, allowing an experimenter to accommodate to the child's execution rhythm. In this early-stage of our research, this paradigm allowed for rapid prototyping of expressive cues without requiring full autonomous robot control.

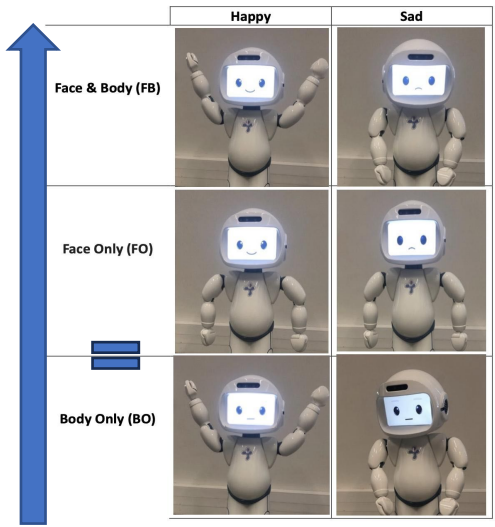

Figure 2. Emotion display and level of embodied expressiveness. From the low to the higher level of nonverbal expressiveness. [BO]: Conveyed emotions through body posture and movement, keeping a neutral facial expression. [FO]: Displayed emotions solely through facial expressions while maintaining a neutral body posture. A smile signified happiness, while a frown indicated sadness. [FB]: Integrated both facial and bodily cues. For happiness, QTRobot smiled and raised its arms; for sadness, it frowned, tilted its head downward, shook its head, and lowered its arms. Cue congruency was respected.

### D. Measure and Data coding

The dependent variable was the child's *alignment* frequency with the *affective feedback* expressed by the robot. Video recordings allowed for retrospective coding of participant *alignment* (or misalignment) for each of the 12 trials. Each trial was assessed based on ball placement. The child was invited to choose a bead. Then:
- If the robot responded by expressing *happiness,* and the child replied by placing the bead in the necklace, the response was coded as "1";
- If the robot responded by expressing *sadness* and the child replaced the previously chosen bead with a new one, the response was coded as "1";
- If the robot responded by expressing *sadness* and the child did not replace his bead with a new one, the response was coded as "0";
- If the robot responded by expressing *happiness* and the child offered a new bead instead, the response was coded as "0".

## IV. RESULTS

The data was analysed using the *lme4* package on *RStudio* (R4.1.1) with a significance level set at .05. Frequency counts of children's response decisions were obtained for each modality and emotion (Table I).

TABLE I. FREQUENCY OF ALIGNMENT AND MISALIGNMENT

| Modality | Happy | | Sad | |
|---|---|---|---|---|
| | Aligned | Misaligned | Aligned | Misaligned |
| Body Only | 39 | 5 | 42 | 2 |
| Body and Face | 38 | 6 | 41 | 3 |
| Face Only | 40 | 4 | 41 | 3 |

To test H1, a GLMM (N=22) was fitted with a binomial distribution and logit link function to examine whether children's alignment with the robot's inferred internal state differed from chance. The dependent variable was *alignment*, with a fixed intercept and a random intercept for participants to account for individual variability. Results showed that the aligned responses were significantly higher than chance (β=3.32, SE=0.60, z=5.58, p<.001), indicating that the overall probability of aligned responses was significantly above chance (≈.97). This suggests that, after controlling for participant-level variability, children tended to align their behavior with the robot's internal state more often than would be expected by random choice.

To test H2 a GLMM (N=22) accounting for both fixed and random effects was run. We specified alignment score as the outcome variable. Modality (FO, BO, FB), emotion (happiness, sadness), trial number (1-12), age, gender order of emotion (1st, 2nd) were included as fixed effects. Random effects accounted for individual differences. The analysis revealed no significant fixed effects for any of the variables (Model 1). However, a tendency (p=.08) was found on emotion, suggesting that in the sad condition, children could more likely change their behavior by selecting a new bead, whereas in the happy condition, where children tend to keep their original choice of bead.

TABLE II. FIXED EFFECTS AND INTERACTIONS

| Variable | Model 1 | | | | | | Model 2 | |
|---|---|---|---|---|---|---|---|---|
| | b | se | z | F | p | 95% CIs | F | p |
| (Intercept) | 0.88 | 1.62 | 0.54 | | 0.59 | [-2.30, 4,06] | | |
| Age | 0.56 | 0.32 | 1.74 | | 0.08 | [-0.07, 1.19] | | |
| Gender | 1.06 | 0.84 | 1.26 | | 0.21 | [-0.59, 2.70] | | |
| Trial | 0.41 | 0.66 | 0.62 | | 0.54 | [-0.89, 1.71] | | |
| Order | -0.03 | 0.09 | -0.32 | | 0.75 | [-0.22, 0.16] | | |
| Emotion | | | | 144.80 | 0.08 | | | |
| Modality | | | | 140.17 | 0.77 | | | |
| Emotion x Modality | | | | | | | 0.38 | 0.84 |
| AIC | 143.64 | | | | | | 147.3 | |

Likewise, the GLMM (Model 2) exploring the interaction between modality and emotion was non-significant, *p*>.05, (Table II). Assessment of model fit revealed the first model without the interaction provided a better fit for the data, ΔAIC=141.4, indicating that the first model was a better predictor of children's ball placement choices during the task. This implies that children in our sample were equally good at recognizing emotion cues presented through isolated cues (facial/bodily) and combined (bodily and facial) cues, and, consequently, at inferring the emotional state (sad or happy) of QTRobot.

## V. DISCUSSION

Our results, in line with previous research with adults (e.g., [23], [26], [27]), show that children tend to align their choices with the robot's inferred internal state when it displays expressions of sadness or happiness, regardless of the modality of expression. Contrary to our predictions, combining facial and bodily cues did not statistically increase alignment (rejecting H2). The absence of a modality effect is likely due to the limited bodily cues provided by QTRobot, which may reduce its ability to convey sufficiently salient or nuanced gestures, leading to reduced expressiveness and potential confusion [5]. However, children responded to the robot's inferred emotional state by confirming or changing their first ball choice to align accordingly (supporting H1).

While we obtained evidence of behavioural alignment our results do not necessarily imply that children's behaviour was motivated by empathic concern. Instead, other social-cognitive processes such as compliance to requests, social conformity, or children's desire to maintain positive interaction dynamics may be at play. With respect to the persuasive value of artificial emotions, similar to studies with adults using nonverbal expressive body and facial cues [28], we observed that emotion expressions influenced children in their willingness to help the robot in the pursuit of a common goal.

However, as in Beran et al. [38], it remains challenging to determine why children adjusted their behavior after receiving negative feedback. One possibility is that they recognise the robot as a social agent and thus perceive helping as the 'right' moral choice [39] in a cooperative setting. Another option is that children simply enjoyed the interaction with the robot and hence were more attentive to catering to its implied requests. Regardless of this, children tends to differentiate between two emotions and interpret the sad expression as a request to modify their first choice and "empathically" align following the same social norms of human cooperation.

Overall, our study cannot allow us to clearly identify the underlying socio-cognitive mechanism underlying children's observed behavior. Regardless of whether children acted out of empathy (perceiving the robot's sadness as a request for support), conformity (aligning with perceived expectations), or simply a desire to please a likable interactive partner, these mechanisms are not mutually exclusive and resonate with the constructs of compliance, empathic concern, and affective influence discussed in the social psychology literature [19], [20].

### A. Limitations and suggestions for future research

A key limitation of this pilot study is the absence of qualitative indicators to help disentangle the motivational mechanisms: whether alignment behavior is driven by empathy, compliance or other factors. To overcome this limitation post-task interviews could be useful to understand what mechanisms are at play and could be employed in future work. Moreover only sadness and happiness were tested as emotional expressions. Exploring a wider range of emotions could offer richer insights into children's empathic and behavioral responses toward robots. Moreover, the study was constrained to only two levels of embodied expressiveness (Face/Body Only and Face and Body). In order to better assess the actual added value of embodied expressiveness in face-to-face CRI, future research will benefit from comparing these nonverbal modalities with a neutral scenario — for instance, where the robot verbally expresses emotions in a neutral tone (e.g., saying "I'm happy" or "I'm sad" without expressive modulation). Finally, the small sample constrains the broader generalizability of our first results.

### B. Implications for shaping our hybrid future with robots

A future where anthropomorphic robots are integrated into our society as social and empathic agents requires a deeper understanding of the psychological mechanisms governing child-robot interactions and the impact of emotions as a means of influence [4], [5]. Rooted in human's sensitivity to attribute human-like characteristics to anthropomorphic robots, our exploratory study, among others, shows that children are notably inclined to follow a robot's suggestions [7]. Children in this context appeared to respond to the robot's emotional expression, showing a sense of shared understanding towards the common goal of the necklace project, highlighting children's tendency to form social and emotional bonds with interactive robots [1]. Non-verbal cues appear sufficient to trigger in children positive *empathic-like* behavior toward them. Whether or not we should allow children to "empathise" with robots, is a deep ethical question that we will not treat here [40] but that definitely needs to be clarified before these systems become pervasive in educational and domestic environments. Critical reflection on the ethical boundaries of emotional persuasion in artificial agents applies not only to AI in general, but also specifically to the ways in which children evolve and grow alongside future empathic robots endowed with AI.

ACKNOWLEDGMENT

The authors wish to acknowledge the *Maison de Jeunes* school, the children who took part in this study as well as their parents. This work was partially funded by *Université Paris 8* [AAP-RECHERCHE ROBODIES – EMO2024] and the *ANR - PEPR O2R AS3* [ANR-22-EXOD-007].